\documentclass{article}
\usepackage{spconf,amsmath,graphicx}
\usepackage[T1]{fontenc}

\usepackage{xcolor}
\usepackage{amsfonts}
\usepackage{tabularx}
\usepackage{amsmath}
\usepackage{graphicx}   
\usepackage{subcaption} 

\usepackage{multirow}  
\usepackage{booktabs} 
\usepackage{url} 
\usepackage{hyperref}

\usepackage[linesnumbered,ruled,vlined]{algorithm2e}

\title{Adaptative Context Normalization:\\ A Boost for Deep Learning in Image Processing}
\name{Bilal FAYE~\textsuperscript{1}, Hanane AZZAG~\textsuperscript{1}, Mustapha LEBBAH~\textsuperscript{2}, Djamel BOUCHAFFRA~\textsuperscript{3}}
\address{\textsuperscript{1}~LIPN, UMR CNRS 7030,
Sorbonne Paris Nord University,
Villetaneuse, France\\ 
\textsuperscript{2}~Paris-Saclay University, UVSQ, David Lab, 78035, Versailles, France.\\ \textsuperscript{3}~Center for Development of Advanced Technologies,
Algiers, Algeria}
\begin{document}
%
\maketitle
\begin{abstract}
Deep Neural network learning for image processing faces major challenges related to changes in distribution across layers, which disrupt model convergence and performance. Activation normalization methods, such as Batch Normalization (BN), have revolutionized this field, but they rely on the simplified assumption that data distribution can be modelled by a single Gaussian distribution. To overcome these limitations, Mixture Normalization (MN) introduced an approach based on a Gaussian Mixture Model (GMM), assuming multiple components to model the data. However, this method entails substantial computational requirements associated with the use of Expectation-Maximization algorithm to estimate parameters of each Gaussian components. To address this issue, we introduce Adaptative Context Normalization (ACN), a novel supervised approach that introduces the concept of "context", which groups together a set of data with similar characteristics. Data belonging to the same context are normalized using the same parameters, enabling local representation based on contexts. For each context, the normalized parameters, as the model weights are learned during the backpropagation phase. ACN not only ensures speed, convergence, and superior performance compared to BN and MN but also presents a fresh perspective that underscores its particular efficacy in the field of image processing. We release our code at \href{https://anonymous.4open.science/r/Adaptative-Context-Normalization-5970}{https://github.com/b-faye/Adaptative-Context-Normalization}.

\end{abstract}
\section{Itroduction}
Normalization, a common data processing operation~\cite{kessy2018optimal}, equalizes variable amplitudes, aiding single-layer network convergence~\cite{lecun2002efficient}. In multilayer networks, data distribution changes necessitate various normalization techniques, such as activation, weight, and gradient normalization. Batch Normalization (BN)~\cite{ioffe2015batch} is a common method for stabilizing multilayer neural network training by standardizing layer activations using batch statistics. While it allows for higher learning rates, BN is limited by batch size dependence and the assumption of uniform data distribution. Specialized variants of BN have been proposed to address batch size-related limitations~\cite{huang2020normalization}. Another approach, Mixture Normalization (MN)~\cite{kalayeh2019training}, was introduced to handle data distribution assumptions. MN differs from BN by accommodating data samples from various distributions. It employs the Expectation-Maximization (EM) algorithm~\cite{Dempster77maximumlikelihood} to cluster data based on components and normalizes each sample using component-specific statistics. MN enhances convergence, particularly in convolutional neural networks, compared to BN.
However, the use of the Expectation-Maximization (EM) algorithm in mixture normalization can lead to a significant increase in computation time, limiting its efficiency. EM estimates the parameters over the entire dataset before the training of the neural network, which enables the provision of normalization parameters.\newline
\indent To tackle this issue, we introduce a novel normalization method called Adaptative Context Normalization (ACN). ACN incorporates a prior knowledge structure called "context", which can be defined as a cluster or class of samples sharing similar characteristics. Building contexts relies on experts' knowledge to group data effectively, but it may result from a partitioning done with another clustering algorithm. For instance, in a dataset with classes and superclasses (groupings of classes), each superclass can be treated as a distinct context. In domain adaptation, each specific domain within the dataset can be considered as independent context. Operating on the hypothesis that activations can be represented as a Gaussian mixture model, ACN normalizes these activations during deep neural network training to estimate parameters for each mixture component. It's worth noting that the choice of a Gaussian distribution assumption is made to ensure a coherent basis for comparison. However, it is important to emphasize that alternative hypotheses for data distribution could also be considered in different scenarios and experiments. As an integral layer within the deep neural network, ACN plays a pivotal role in standardizing activations originating from the same context using the parameters acquired through backpropagation. This process facilitates the estimation of parameters for each mixture component, ultimately enhancing the data representation's discriminative capacity with respect to the target task.\newline 
This study introduces three significant contributions:
\begin{itemize}
    \item We introduce the concept of "context", formed by expert knowledge similar to superclasses or specific scenarios in image classification. Our innovative method, \textbf{ACN}, acts as a layer in the neural network. ACN supervises and estimates normalization parameters specific to Gaussian mixture components according predefined contexts.
    
    \item  We apply ACN to various deep neural network architectures for image processing using Vision Transformer and Convolutional Neural Networks, serving as a layer at different levels. Extensive experiments consistently show that ACN accelerates training processes, improving generalization performance.
    
    \item We build upon ACN's strong image classification performance, extending its application to image domain adaptation. This aims to enhance model performance across diverse data distributions, effectively addressing the associated challenges.
\end{itemize}

\section{Related Work}
\subsection{Batch Normalization}\label{section:batch_normalization}
To ensure clear understanding and consistency in our model, and to facilitate comparison with existing work, we adopt the same notations as those used in~\cite{kalayeh2019training}. Let's consider $x \in \mathbb R^{N\times C\times H\times W}$, a 4-D  activation tensor in a convolutional neural network (CNN), where the axes $N$, $C$, $H$, and $W$ represent the batch size, number of channels, height, and width respectively. Batch normalization (BN) is performed on the mini-batch $B = \{x_{1:m}: m \in [1, N]\times [1, H] \times [1, W]\}$, where $x$ is flattened across all but channel:
\begin{equation}
    \hat{x}_{i} = \frac{x_{i}-\mu_B}{\sqrt{\sigma^2_B+\epsilon}},
    \label{bn_equation}
\end{equation}
where $\mu_B = \frac{1}{m} \sum_{i=1}^m x_{i}$ and $\sigma^2_B = \frac{1}{m}\sum_{i=1}^m (x_{i}-\mu_B)^2$ represent, respectively, the mean and variance of $B$, while $\epsilon > 0$ is a small value that handles numerical instabilities.
If samples within the mini-batch are from the same distribution, the transformation shown in Equation~\ref{bn_equation} generates a zero mean and unit variance distribution. This constraint allows stabilizing the distribution of the activations and thus benefits training.

\subsection{Variants of Batch Normalization}\label{section:mn}
Several extensions of BN~\cite{huang2020normalization} have been proposed, including Layer Normalization (LN). LN is tailored for recurrent neural networks and aims to eliminate dependencies among activations by normalizing each neuron activation with respect to its respective layer. 
Another technique is Instance Normalization (IN), which normalizes each sample individually, focusing on removing style information, especially in images. IN improves the performance of specific deep neural networks and finds widespread application in tasks like image
style transfer.
Similarly, Group Normalization (GN), 
divides neuron activations into groups and independently normalizes activations in
these groups. Like IN, GN is effective for visual tasks with small batch sizes, such as object
detection and segmentation. Lastly, Mixture Normalization (MN)~\cite{kalayeh2019training}, which uses a Gaussian Mixture Model (GMM) to normalize activations based
on multiple modes of variation in their underlying distribution, suitable when
the distribution of activations exhibits multiple modes.
The general transformation $x \rightarrow \hat{x}$ is shared across all these variants and is applied on the flattened $x$ along the spatial dimension $L = H \times W$, as follows:
\begin{equation}
    v_{i} = x_{i} - \mathbb{E}_{B_i}(x), \ 
    \hat{x}_{i} = \frac{v_{i}}{\sqrt{\mathbb{E}_{B_i}(v^2)+\epsilon}},
    \label{general_transform}
\end{equation}
where $B_i = \{j: j_N \in [1, N], j_C \in [i_C], j_L \in [1, L]\}$,
and $i = (i_N, i_C, i_L)$ a vector indexing the activations $x \in \mathbb R^{N\times C \times L}$.\newline
 In MN algorithm, each $x_i$ in $B_i$ is normalized with the mean and standard deviation of the mixture component it belongs to. The probability density function $p_\theta$ can be represented as a GMM.
Let $x \in \mathbb R^D$, and $\theta = \{\lambda_k, \mu_k, \Sigma_k: k = 1, ..., K\}$, then we have:
  $$  p(x) = \sum_{k=1}^K \lambda_k p(x|k),\ \text{s.t.}\ \forall_k\ :\ \lambda_k\ \ge 0,\ \sum_{k=1}^{K}\lambda_k=1,$$
where
 $$   p(x|k) = \frac{1}{(2\pi)^{D/2}\lvert\Sigma_k\rvert^{1/2}}\exp\left(-\frac{( x-\mu_k)^T\Sigma_k^{-1}( x-\mu_k)}{2}\right),$$
is the $k$-th component, $\mu_k$ is the mean vector, and $\Sigma_k$ is the covariance matrix.
The probability that $x$ was generated by the $k$-th Gaussian component can be defined as follows:
 $$   \tau_k(x) = p(k|x) = \frac{\lambda_k p(x|k)}{\sum_{j=1}^K\lambda_j p(x|j)}$$
Based on these assumptions and the general transformation in Equation~\eqref{general_transform}, the normalization of $x_i$ is defined as follows:
\begin{equation}
    \label{mn_aggregation}
    \hat{x}_{i} = \sum_{k=1}^K \frac{\tau_k(x_{i})}{\sqrt{\lambda_k}}\hat{x}_{i}^k,
\end{equation}
with
\begin{equation}
    \label{mn_norm}
    v_{i}^k=x_{i} - \mathbb E_{B_i}[\hat{\tau}_k(x).x], \
    \hat{x}_{i}^k = \frac{v_{i}^k}{\sqrt{\mathbb E_{B_i}[\hat{\tau}_k(x).(v^k)^2]+\epsilon}},
\end{equation}
where
 $   \hat{\tau}_k(x_{i}) = \frac{\tau_k(x_{i})}{\sum_{j \in B_i} \tau_k(x_{j})},$
is the normalized contribution of $x_i$ in estimating the statistics of the $k$-th Gaussian component.
With this approach, MN can be applied in two steps:
\begin{enumerate}
\item Estimation of the mixture model parameters 
$\theta $
using the EM algorithm~\cite{Dempster77maximumlikelihood}.
\item Normalization of each $x_i$ with respect to the estimated parameters (Equations\eqref{mn_norm},\eqref{mn_aggregation}) 
\end{enumerate}
\section{Proposed Method: Adaptative Context Normalization}\label{section:cn}
In this framework, clusters of samples sharing common characteristics and behaviors are defined by the "context". Each context is characterized by a unique input scalar ($r$), and activations originating from samples within a specific context undergo normalization. This normalization process is facilitated through parameters $\theta_r = \{\mu_r, \sigma_r\}$, which are acquired through neural network backpropagation. To ensure coherence, we adhere to the assumption of Gaussian mixture distributions in activations, aligning with the predefined understanding of contexts. It is crucial to note that various experiments may explore alternative hypotheses regarding data distribution or normalization methods.
\newline
Let $x$ be an activation tensor in the $\mathbb{R}^{N \times C \times H \times W}$ space, where $B_i$ represents a group of activations resulting from flattening $x$ along the $L = H \times W$ axis. Each $x_i$ within $B_i$ is normalized along the $C$ axis, using the parameters associated with its context, $\theta_{r_i} = \{\mu_{r_i}, \sigma_{r_i}\}$:
\begin{equation}\label{equation:cn_nom}
    ACN_{\theta_{r_i}}: \hat{x}_i \gets \frac{x_i - \mu_{r_i}}{\sqrt{\sigma^2_{r_i}+\epsilon}}
\end{equation}
The parameters $\theta = \{\mu_r, \sigma_r\}_{r=1}^T$, learned for each of the $T$ mixture components, undergo updates during the normalization process applied to the activations from the corresponding context.
Each normalized activation $\hat{x}_i$
can be viewed as an input to a sub-network composed of the linear transform (Equation~\ref{equation:cn_nom}), followed by the other processing done by the original network.
\begin{algorithm}[t]
\caption{ Training an Adaptative Context-Normalized Network}\label{alg:one}
\SetKwInOut{KwIn}{Input}
\SetKwInOut{KwOut}{Output}

\KwIn{Deep neural network $Net$ with trainable parameters $\Theta$; subset of activations and its contexts $\{x_i, r_i\}_{i=1}^m$, with $r_i \in \{1, ..., T\}$, where $T$ is the number of contexts
}
\KwOut{Adaptative Context-Normalized deep neural network for inference, $Net^{inf}_{ACN}$}
    $Net^{tr}_{ACN} = Net$ // {\small \it Training ACN deep neural network }\\ 
    \For{$i \gets 1$ to $m$}{
        \begin{itemize}
            \item Add transformation $\hat{x}_i$ = $ACN_{\theta_{r_i}}$($x_i$) to $Net^{tr}_{ACN}$ (Equation~\ref{equation:cn_nom})
            \item Modify each layer in $Net^{tr}_{ACN}$ with input $x_i$ to take $\hat{x}_i$ instead
        \end{itemize}
    }
    Train $Net^{tr}_{ACN}$ to optimize the parameters: $\Theta = \Theta \cup \{\mu_r, \sigma_r\}_{r=1}^T$ \\ 
    $Net^{inf}_{ACN} = Net^{tr}_{ACN}$ // {\small \it Inference ACN network with frozen parameters}\\ 
    \For{$i \gets 1$ to $m$}{
            \begin{itemize}
                \item Retrieve the parameters associated with the context of $x_i$: $\theta_{r_i} = \{\mu_{r_i}, \sigma_{r_i}\} $ 
                \item transform $\hat{x}_i$ = $ACN_{\theta_{r_i}}$($x_i$) using Equation~\ref{equation:cn_nom}
            \end{itemize}
    }
\end{algorithm}
During the training process, as outlined in Algorithm~\ref{alg:one}, it is necessary to propagate the gradient of the loss function $\ell$, through the transformation. Furthermore, it is essential to calculate the gradients with respect to the parameters of the ACN transform. This computation is accomplished using the chain rule, as illustrated in the following expression (before simplification):
$$ \frac{\partial \ell}{\partial \mu_{r_i}} = \frac{\partial \ell}{\partial \hat{x}_i}.\frac{\partial \hat{x}_i}{\partial \mu_{r_i}} = -\frac{\partial \ell}{\partial \hat{x}_i}.(\sigma_{r_i}^2+\epsilon)^{-1/2}$$
$$\frac{\partial \ell}{\partial \sigma_{r_i}^2} = \frac{\partial \ell}{\partial \hat{x}_i}.\frac{\partial \hat{x}_i}{\partial \sigma_{r_i}^2} = \frac{\mu_{r_i} + x_i}{2(\sigma_{r_i}^2 + \epsilon)^{3/2}}$$
ACN normalizes neural network activations, aiding convergence and creating Gaussian components for task-specific representations in a latent space.\newline
\indent During inference, we can either normalize activations based on their corresponding context as shown in Algorithm~\ref{alg:one} or treat all contexts collectively (as mixture normalization), by transforming Equation~\ref{mn_aggregation}, as follows:
\begin{equation}
    \label{cn_aggregation}
     \hat{x}_{i} = \sqrt{T}\sum_{r=1}^{T} \tau_{r}(x_{i})\hat{x}_{i}^{r} ,
\end{equation}
with
 $   v_{i}^{r}=x_{i} - \mathbb E_{B_i}[\hat{\tau}_r(x).x], \ 
    \hat{x}_{i}^{r} = \frac{v_{i}^{r}}{\sqrt{\mathbb E_{B_i}[\hat{\tau}_{r}(x).(v^{r})^2]+\epsilon}},$
where
 $   \hat{\tau}_{r}(x_{i}) = \frac{\tau_{r}(x_{i})}{\sum_{j \in B_i} \tau_{r}(x_{j})},$
In Equation~\ref{cn_aggregation}, $T$ represents the number of contexts, and we assume constant prior probabilities ($\lambda_{r} = \frac{1}{T}, {r}=1, ..., T$).
\section{Experiments}
In the proposed experiments, we will assess the performance of the adaptative context normalization method ACN and compare it to BN and MN. We also evaluate our results against ACN-base, a simplified variant of ACN. In ACN-base, two separate neural networks estimate the $T$ context parameters $\theta=\{\mu_r, \sigma_r\}_{r=1}^T$. For a given input $x_i$ in the ACN-base layer, the context identifier $r_i$ of $x_i$ is initially encoded using one-hot encoding and serves as input for both networks. The first neural network outputs $\mu_{r_i}$, while the second provides $\sigma^2_{r_i}$, which are then used to normalize $x_i$.
The experiments will cover different architectural setups, including Convolutional Neural Networks (CNNs) (as described in Sections~\ref{section:cn1}, and \ref{section:cn3}) and Vision Transformers (ViT)~\cite{dosovitskiy2020image} (as outlined in Section~\ref{section:vit}). These approaches will be evaluated across a range of tasks, including classification and domain adaptation.
%
\subsection{Datasets}\label{section:datasets}
The experiments conducted in this study utilize several commonly used benchmark datasets in the classification community, including:
\begin{itemize}
        \item \textbf{CIFAR-10 and CIFAR-100:} These datasets include 50,000 training images and 10,000 test images, each sized $32\times32$ pixels. CIFAR-10 features 10 distinct classes, while CIFAR-100 has 100 classes organized into 20 superclasses~\cite{cifar_datasets}.
        \item \textbf{Tiny ImageNet:} A dataset that is a reduced version of the ImageNet dataset, containing 200 classes with 500 training images and 50 test images per class~\cite{le2015tiny}.
        \item \textbf{MNIST digits:} Contains 70,000 grayscale images of size $28\times28$ pixels representing the 10 digits, with approximately 6,000 training images and 1,000 test images per class~\cite{mnist_datasets}.
        \item \textbf{SVHN:} A dataset with over 600,000 digit images, focused on digit and number recognition in natural scene images~\cite{sermanet2012convolutional}.
\end{itemize}
These datasets provide a diverse set of challenges for evaluating the proposed approaches in various tasks.
\subsection{A Comparative Study: Adaptative Context Normalization vs. Mixture Normalization 
}\label{section:cn1}
In this experiment, we employ a shallow deep Convolutional Neural Network (ConvNet) architecture as described in the MN paper. This network consists of four convolutional layers with ReLU activation, each followed by a batch normalization layer. One challenge associated with batch normalization (BN) is the utilization of non-linear functions (e.g., ReLU) subsequent to activation normalization. Through the application of ConvNet, we showcase how Adaptative Context Normalization (ACN) addresses this issue and improves convergence and overall performance by substituting a BN layer with ACN within the ConvNet.


\indent During training on CIFAR-10, CIFAR-100, and Tiny ImageNet datasets, we employ the MN method, which involves estimating a Gaussian mixture model through Maximum Likelihood Estimation (MLE). For comparison purposes, the three components discovered by the Expectation-Maximization (EM) algorithm on MN are utilized as distinct contexts ($T=3$) for ACN and ACN-base, enabling the normalization of samples within each context. We vary the learning rate from 0.001 to 0.005, use a batch size of 256, and train for 100 epochs with the AdamW optimizer~\cite{loshchilov2017decoupled,kingma2014adam}. To compare normalization methods, we replace the third BN layer in ConvNet with an MN layer and repeat this process for ACN and ACN-base layers.

\begin{figure*}[h]
    \centering
    \begin{subfigure}[b]{0.23\textwidth}
        \centering
        \includegraphics[width=\textwidth]{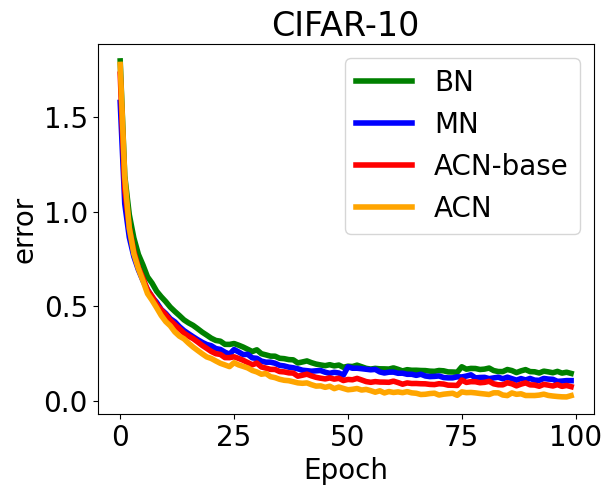}
        \caption{CIFAR-10}
        \label{fig:bn}
    \end{subfigure}
    \hfill
    \begin{subfigure}[b]{0.23\textwidth}
        \centering
        \includegraphics[width=\textwidth]{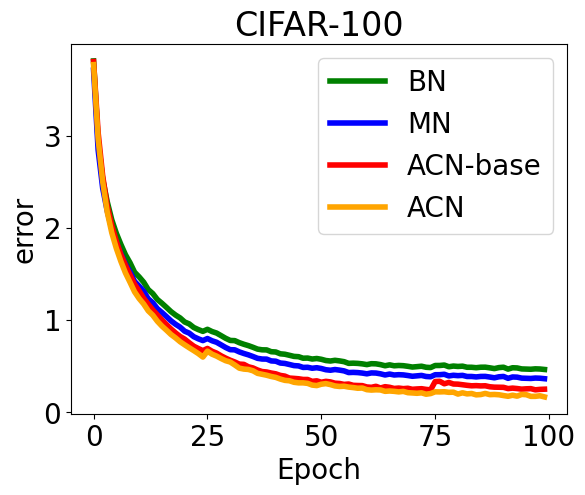}
        \caption{CIFAR-100}
        \label{fig:CN-Patches}
    \end{subfigure}
    \hfill
    \begin{subfigure}[b]{0.23\textwidth}
        \centering
        \includegraphics[width=\textwidth]{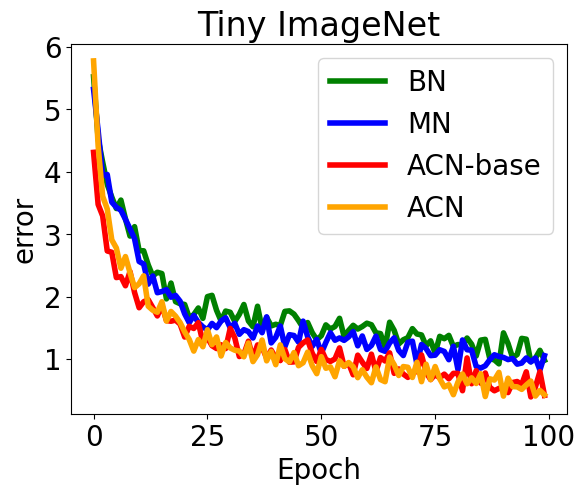}
        \caption{Tiny ImageNet}
        \label{fig:CN-Channels}
    \end{subfigure}
    \caption*{Learning rate = 0.001}
    \label{fig:cifar100_loss}


    \begin{subfigure}[b]{0.23\textwidth}
        \centering
        \includegraphics[width=\textwidth]{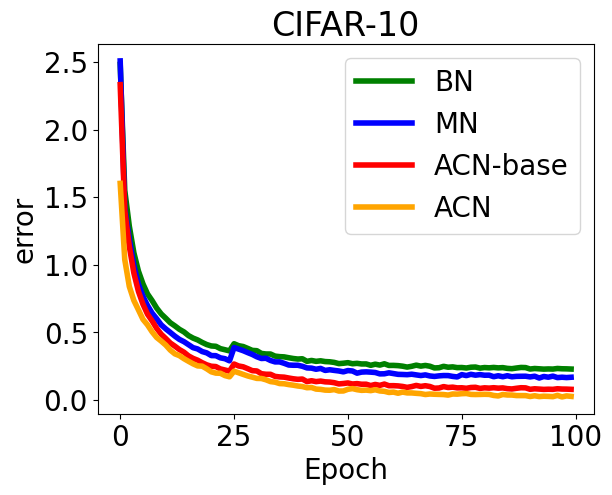}
        \caption{CIFAR-10}
        \label{fig:figure1}
    \end{subfigure}
    \hfill
    \begin{subfigure}[b]{0.23\textwidth}
        \centering
        \includegraphics[width=\textwidth]{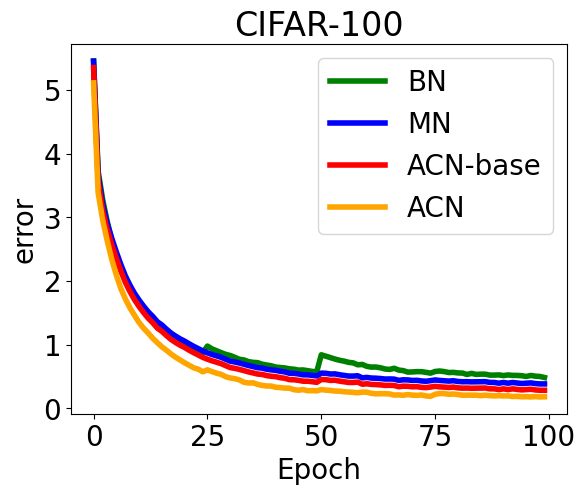}
        \caption{CIFAR-100}
        \label{fig:figure2}
    \end{subfigure}
    \hfill
    \begin{subfigure}[b]{0.23\textwidth}
        \centering
        \includegraphics[width=\textwidth]{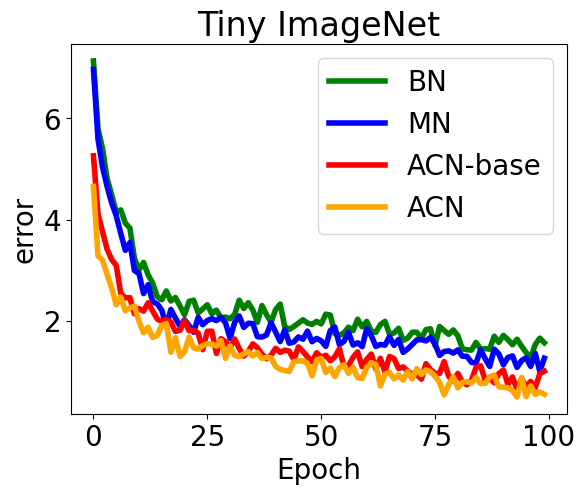}
        \caption{Tiny ImageNet}
        \label{fig:figure3}
    \end{subfigure}
    \caption*{Learning rate = 0.005}
    \label{fig:second_set_of_figures}
    \caption{Test error curves when ConvNet architecture is trained under different learning rate.}
    \label{fig:all_figures}
\end{figure*}

Figure~\ref{fig:all_figures} demonstrates that ACN and its smaller variant, ACN-base, exhibit faster convergence compared to BN and MN. This accelerated convergence results in improved performance on the validation dataset, with an average increase of \textbf{2\%} in accuracy on CIFAR-10, \textbf{3\%} on CIFAR-100, and \textbf{4\%} on Tiny ImageNet. This positive trend persists across different numbers of classes (10, 100, 200), even with an increase in the learning rate from 0.001 to 0.005. Increasing the learning rate exacerbates the gap in convergence, demonstrating the capacity of our normalization technique to effectively leverage higher learning rates during training.\newline
\indent In our upcoming experiments, we aim to demonstrate that adaptative context normalization can be implemented in a single step for specific scenarios, offering a potential reduction in time complexity compared to the more intricate mixture normalization. However, it's worth noting that in the experiments described in the following sections, we do not utilize Gaussian components as context, rendering MN inapplicable as a baseline for comparison.

\subsection{Using Superclasses as Contexts in Adaptative Normalization
}
\label{section:vit}
In our adaptative context normalization approach, superclasses serve as prior knowledge. Each superclass represents a distinct context. ACN is integrated into the Vision Transformer (ViT) for CIFAR-100 classification using Keras baseline~\cite{keras_vit}. 
The core innovation in the experiment lies in the use of CIFAR-100 superclasses as contexts for predicting the dataset's 100 classes, specifically in the case of ACN method.

\begin{figure}[h]
\centering
\includegraphics[width=8.6cm]{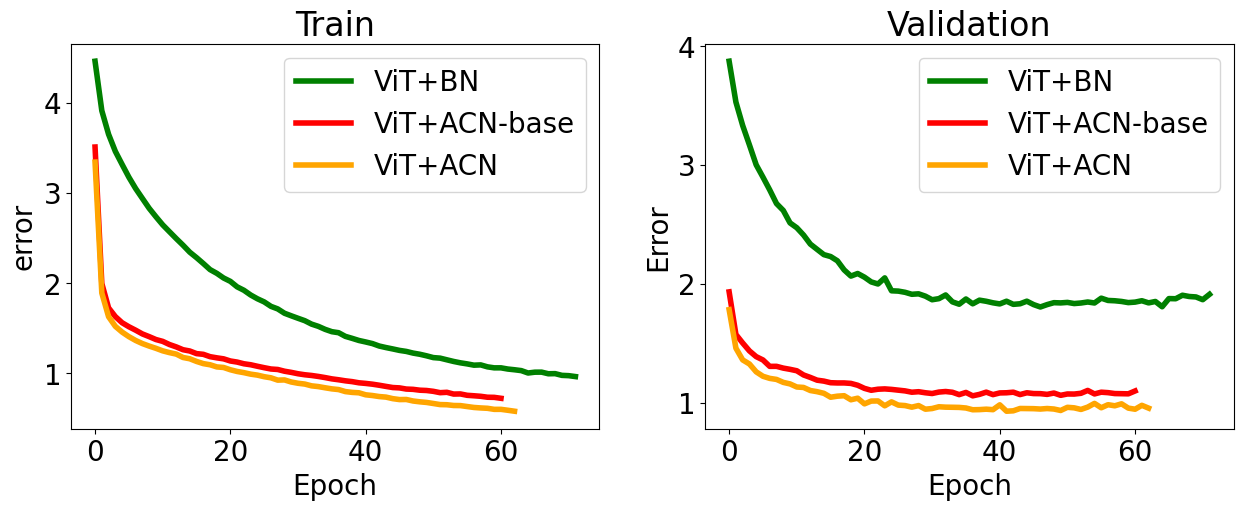}
\caption{Comparing Training and Validation Error Curves: ACN-base and ACN in ViT Architecture on CIFAR-100 show faster convergence and lower validation loss, enhancing learning efficiency and classification compared to BN.}
\label{figure:cifar100_superclass}
\end{figure}

We employed three distinct models: the base ViT model sourced from Keras~\cite{keras_vit}, a modified version incorporating a Batch Normalization (BN) layer as its initial component, and two alternative models that replaced the BN layer with ACN-base and ACN layers. Training employed early stopping based on validation performance, and images were pre-processed by normalizing them with respect to the dataset's mean and standard deviation. Data augmentation techniques such as horizontal flipping and random cropping were applied to enhance the dataset. The AdamW optimizer, with a learning rate of $10^{-3}$ and a weight decay of $10^{-4}$, was selected to prevent overfitting and optimize model parameters~\cite{loshchilov2017decoupled,kingma2014adam}.
\begin{table}[t]
    \centering
    \begin{tabular}{llllll}
        \hline
        model & accuracy & precision & recall & f1-score \\
        \hline
        ViT+BN & 55.63 &  8.96 & 90.09  & 54.24 \\
        ViT+ACN-base & 65.87 & \textbf{23.36} & 98.53 & 65.69 \\
        ViT+ACN & \textbf{67.38} & \textbf{22.93} & \textbf{99.00} & \textbf{67.13} \\
        \hline
    \end{tabular}
    \caption{Evaluating CIFAR-100 Performance with ViT Architecture~\cite{keras_vit} integrating BN, ACN-base, and ACN with superclasses as contexts.}
    \label{table:cifar_superclass}
\end{table}
Table~\ref{table:cifar_superclass} showcases the substantial performance improvements achieved by our novel Adaptative Context Normalization (ACN), when compared to Batch Normalization (BN) training the ViT architecture from scratch on CIFAR-100. ACN approach exhibit an accuracy improvement of approximately \textbf{12\%} over BN. Additionally, it's worth noting that ACN-base and ACN demonstrate faster convergence compared to BN, requiring less training time to achieve superior performance. This observation is further supported by the train and validation loss comparison depicted in Figure~\ref{figure:cifar100_superclass}, highlighting that ACN facilitates accelerated learning while enhancing classification performance. These findings suggest that the proposed method ACN not only stabilizes data distributions and mitigates internal covariate shift but also significantly reduces training time for improved results. ViT+ACN-base and ViT+ACN achieve outstanding performance, surpassing all known ViT models when trained from scratch on the CIFAR-100 dataset.
\vspace{-0.5cm}

\subsection{Domain Adaptation: Using Source and Target Domains as Contexts}\label{section:cn3}
\begin{table}[!t]
    \centering
    \begin{tabular}{llllll}
        \hline
        \multicolumn{5}{c}{\textbf{MNIST (source domain)}}\\
        \hline
        model  & accuracy & precision & recall & f1-score  \\
        \hline
        AdaMatch & 97.36 & 87.33 & 79.39 & 78.09 \\
        AdaMatch+ACN-base & \textbf{99.26} & \textbf{99.20} & \textbf{99.32} & \textbf{99.26} \\
        \hline
        AdaMatch+ACN & 98.92 & 98.93 & 98.92 & 98.92\\
        \hline
    \end{tabular}

    \begin{tabular}{llllll}
        \hline
        \multicolumn{5}{c}{\textbf{SVHN (target domain)}}\\
        \hline
        model  & accuracy & precision & recall & f1-score  \\
        \hline
        AdaMatch & 25.08 & 31.64 & 20.46 & 24.73 \\
        AdaMatch+ACN-base & 43.10 & 53.83 & 43.10 &  47.46 \\
        \hline
        AdaMatch+ACN & \textbf{54.70} & \textbf{59.74} & \textbf{54.70} & \textbf{54.55} \\
        \hline
    \end{tabular}
     \caption{Comparing model performance: AdaMatch vs. AdaMatch+ACN-base and AdaMatch+ACN on MNIST (source) and SVHN (target) datasets.}
    \label{table:adamatch}
\end{table}

In this experiment, we demonstrate that adaptative context normalization's (ACN) proficiency in enhancing local representations can lead to significant improvements in domain adaptation. Domain adaptation, as explained in~\cite{farahani2021brief}, involves leveraging knowledge acquired by a model from a related domain, where there is sufficient labeled data, to enhance the model's performance in a target domain with limited labeled data. In this scenario, we consider two contexts: the "source domain" and the "target domain".
%
To illustrate this, we apply ACN in conjunction with AdaMatch~\cite{berthelot2021adamatch}, a method that combines the tasks of unsupervised domain adaptation (UDA), semi-supervised learning (SSL), and semi-supervised domain adaptation (SSDA). In UDA, we have access to a labeled dataset from the source domain and an unlabeled dataset from the target domain. The goal is to train a model that can effectively generalize to the target dataset. It's worth noting that the source and target datasets exhibit variations in distribution. Specifically, we utilize the MNIST dataset as the source dataset, while the target dataset is SVHN. Both datasets encompass various factors of variation, including texture, viewpoint, appearance, etc., and their domains, or distributions, are distinct from each other.\newline

\indent A model (AdaMatch)~\cite{keras_adamatch} trained from scratch with wide residual networks~\cite{zagoruyko2016wide} on dataset pairs, serves as the reference model. The model is trained using the Adam~\cite{kingma2014adam} optimizer and a cosine decay schedule to decrease the initial learning rate, which is initialized at 0.03. ACN is used as initial layer in AdaMatch to incorporate the context identifier (source domain and target domain) into the image normalization process. As the labels in the source domain are known, the model provides a better representation of this domain compared to the target domain, where the labels are unknown. This advantage is leveraged by considering the context of the source domain during inference to enhance model's performance on the target domain.\newline
\indent It is clear that in a broader context, Table~\ref{table:adamatch} demonstrates a significant improvement in validation metrics with the use of adaptative context normalization. This improvement is evident through an increase in accuracy of \textbf{18.02\%} with ACN-base and \textbf{29.62\%} with ACN. This enhancement notably bolsters the performance of the AdaMatch model, resulting in significantly accelerated convergence during training.
\begin{figure}[!t]
  \centering
  \begin{subfigure}{0.45\textwidth}
    \includegraphics[width=\linewidth]{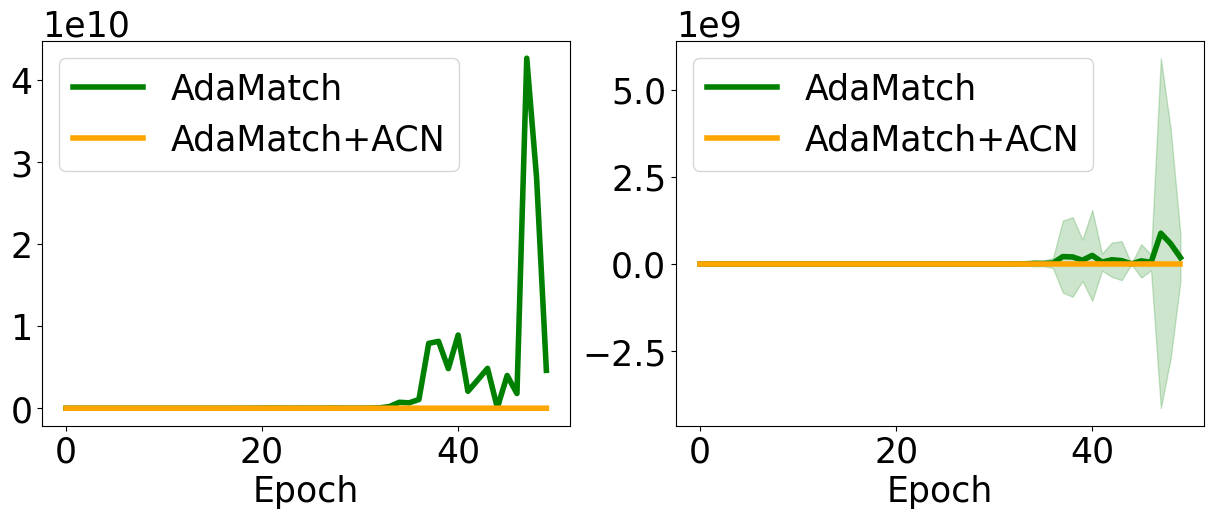}
    \caption{Source domain (MNIST)}
  \end{subfigure}
  \hspace{15pt}
  \begin{subfigure}{0.45\textwidth}
    \includegraphics[width=\linewidth]{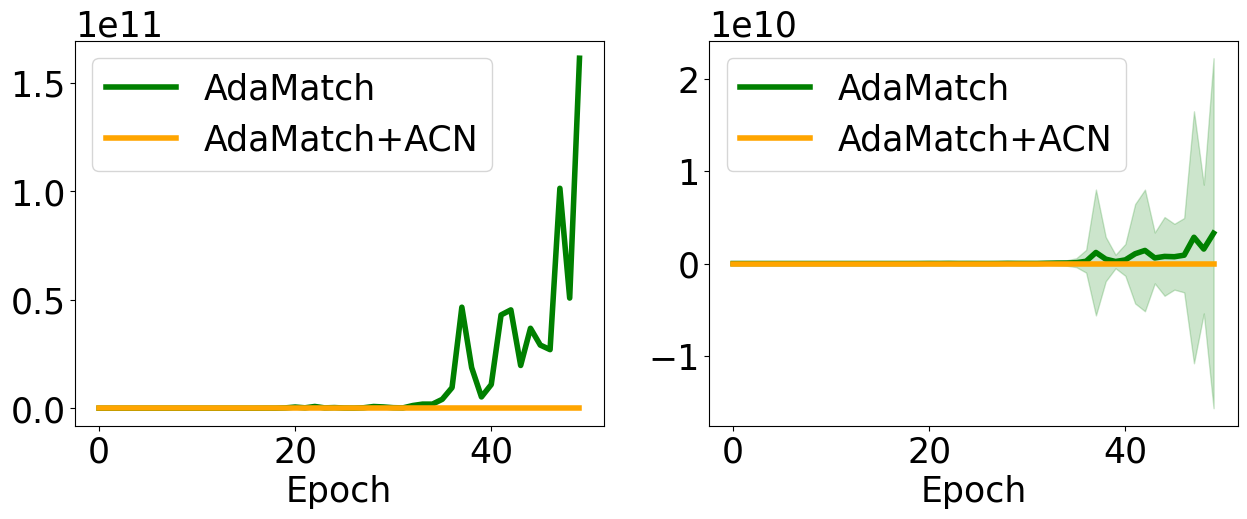}
    \caption{Target domain (SVHN)}
  \end{subfigure}
  \caption{Gradient Variance Evolution: AdaMatch and AdaMatch+ACN models during training on source (MNIST) and target (SVHN) domains. Left: Max gradient variance per epoch. Right: Average gradient variance per epoch.}
  \label{fig:adamatch_gradient}
\end{figure}
Figure~\ref{fig:adamatch_gradient} shows the valuable role that adaptative context normalization plays in stabilizing the gradient throughout the training process, benefiting both the source and target domains. As a result, this contributes to faster convergence and an overall improvement in model performance.

\section{Conclusion}
We present Adaptive Context Normalization (ACN) as a method to improve deep neural network training, with a particular focus on applications in image processing. ACN boosts stability, speeds up convergence, supports higher learning rates, and accommodates various activation functions. Unlike Mixture Normalization, ACN uses a context-based approach, enabling supervised learning of Gaussian component parameters for faster estimation. ACN offers non-linear decision boundaries. 
Through experiments, it has demonstrated superior performance when compared to established methods like batch normalization and mixture normalization, proving its value in domains like domain adaptation and image classification.\newline
\indent Our aim is to create an unsupervised variant of ACN that dynamically acquires context during training, bypassing the need for it as an input. This involves cluster discovery through normalization in training, potentially enhancing deep neural network convergence and performance by customizing data representation to the target task.
\section{Acknowledgements} We thank Grid5000 for providing the computational resources that enabled us to conduct experiments within the framework of the LabCom partnership.
\bibliographystyle{IEEEbib}
\bibliography{strings,refs}

\end{document}